\documentclass[letterpaper, 10 pt, conference]{ieeeconf}  

\IEEEoverridecommandlockouts                              

\usepackage{tikz}
\usepackage{bm}
\usepackage{graphics}
\usepackage{color}
\usepackage{float}
\usepackage{epsf}
\usepackage{subcaption}
\usepackage[font=small]{caption}
\usepackage{xfrac} 
\usepackage{float} 

\usepackage{csquotes}
\usepackage{hyperref}
\usepackage{todonotes}
\usepackage[l3]{csvsimple}
\usepackage{array}
\usepackage{graphicx}    
\usepackage{subcaption}  
\usepackage{lipsum}      
\usepackage{makecell} 
\usepackage{multirow}
\usepackage{booktabs} 
\usepackage{placeins} 
\usepackage[style=ieee,maxnames=2,maxcitenames=2,mincitenames=1,doi=false,isbn=false,url=false,eprint=false, maxbibnames=2, minbibnames=1]{biblatex}
\usepackage{amssymb}
\usepackage{pifont}
\title{\LARGE \bf
The Role of Embodiment in\\ Intuitive Whole-Body Teleoperation for Mobile Manipulation
}

\author{Sophia Bianchi Moyen$^{* 1,2}$, Rickmer Krohn$^{* 1}$, Sophie Lueth$^{* 1}$, \\ Kay Pompetzki$^{1}$, Jan Peters$^{1,3,4,5}$, Vignesh Prasad$^{1}$ and Georgia Chalvatzaki$^{1,3}$
\thanks{$^{*}$Equal contribution; $^{1}$Computer Science Department, TU Darmstadt, Germany; $^{2}$ University of São Paulo, Brazil; $^{3} $Hessian.AI, Darmstadt, Germany; $^{4}$ Centre for Cognitive Science, TU Darmstadt, Germany; $^{5}$ Systems AI for Robot Learning, German Research Center for AI (DFKI). This research is funded by the German Research Foundation (DFG) Emmy Noether Programme (CH 2676/1-1), the EU’s Horizon Europe project \enquote{ARISE} (Grant no.: 101135959) and the German Federal Ministry of Education and Research (BMBF) project Robotics Institute Germany (RiG) (Grant no.: 16ME1001). Email: \texttt{rickmer.krohn@tu-darmstadt.de}}%
}

\begin{document}

\maketitle
\thispagestyle{empty}
\pagestyle{empty}


\begin{abstract}
Intuitive Teleoperation interfaces are essential for mobile manipulation robots to ensure high quality data collection while reducing operator workload. A strong sense of embodiment combined with minimal physical and cognitive demands not only enhances the user experience during large-scale data collection, but also helps maintain data quality over extended periods. This becomes especially crucial for challenging long-horizon mobile manipulation tasks that require whole-body coordination. 
We compare two distinct robot control paradigms: a coupled embodiment integrating arm manipulation and base navigation functions, and a decoupled embodiment treating these systems as separate control entities. Additionally, we evaluate two visual feedback mechanisms: immersive virtual reality and conventional screen-based visualization of the robot's field of view. These configurations were systematically assessed across a complex, multi-stage task sequence requiring integrated planning and execution.
Our results show that the use of VR as a feedback modality increases task completion time, cognitive workload, and perceived effort of the teleoperator. Coupling manipulation and navigation leads to a comparable workload on the user as decoupling the embodiments, while preliminary experiments suggest that data acquired by coupled teleoperation leads to better imitation learning performance.
Our holistic view on intuitive teleoperation interfaces provides valuable insight into collecting high-quality, high-dimensional mobile manipulation data at scale with the human operator in mind. Project website: \url{https://sophiamoyen.github.io/role-embodiment-wbc-moma-teleop/}

\end{abstract}

\section{Introduction}
The availability of large-scale robotic manipulation datasets has increased significantly in recent years, fueling advancements in learning-based approaches for robotic control
~\cite{khazatsky2024droid,
Dasari2019,
walke2023bridgedata,
openx,
sharma2018multipleinteractionseasymime,
}. 
These datasets predominantly focus on stationary robotic arms and rely on teleoperation interfaces that are well-suited for fixed-base manipulators. This reduces the operational complexity to a predetermined, stable workspace, significantly simplifying the control paradigm. 
In contrast, real-world environments, such as households and assistive scenarios, demand mobile manipulators capable of navigating through an environment for executing diverse and robust manipulation policies. Despite the increasing need for such robots, large-scale datasets for mobile manipulation remain limited, with only recent efforts beginning to emerge~\cite{fu2024mobilealohalearningbimanual}. 
Mobility expands the robot’s operational workspace but increases control and feedback complexity. As a result, data collection for mobile manipulation teleoperation becomes more challenging, requiring user to maintain situational awareness over a dynamic and large operation  space, raising cognitive load and the need for effective feedback. Thus, intuitive teleoperation interfaces that balance embodiment, cognitive demand, and task efficiency are essential for scalable, high-quality data collection in mobile manipulation learning.

\begin{figure}[t!]
    \centering    
    \includegraphics[width=\linewidth]{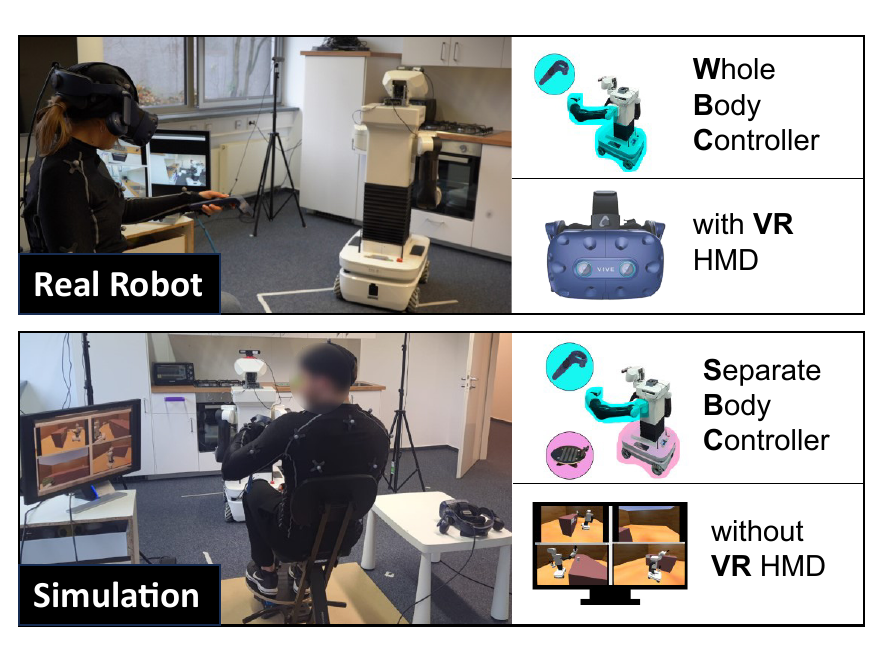}
    \caption{Mobile manipulation teleoperation in Simulation and on the real robot using different Controllers (WBC  and SBC) and visualization modalities (with VR and without).
    }
    \label{fig:teaser}
    \vspace{-0.5cm}
\end{figure}
 
Various control strategies have been proposed to enhance operator efficiency in a task-specific way~\cite{schwarz2017nimbro, nakanishi2020, arduengo2021human, 
}. 
Studies have explored a variety of attitudinal measures, such as workload, usability, simulation sickness, and behavioral metrics, e.g., task completion time, trajectory smoothness, and ergonomic data. In an effort to create a standardized evaluation protocol for mobile manipulation teleoperation, Wan et al.~\cite{Wun2023} proposed a performance and usability evaluation scheme. However, existing studies focus on short-horizon tasks that require minimal manipulation skills, overlooking the complexities inherent in long-horizon mobile manipulation. In real-world scenarios, effective teleoperation involves the precise control and coordination of upper and lower body movements, error recovery, and sustained user experience over extended operation periods. Addressing these challenges is crucial to the design of intuitive and efficient teleoperation interfaces for mobile manipulation in diverse environments.

\hfill

To address these gaps, we present a comprehensive study that goes beyond short-horizon tasks and low-manipulability scenarios by evaluating teleoperation in complex, long-horizon mobile manipulation settings. Our work studies the interplay between control strategies of the robot and feedback mechanisms for the teleoperator. Our main contribution is a holistic analysis of the two key aspects of data collection for mobile manipulation: the teleoperation framework and the feedback interface. Two example combinations can be seen in Fig.~\ref{fig:teaser}. Beyond standard performance metrics, we assess operator experience across extended task durations. Specifically, we examine task performance, physical and cognitive workloads to provide insights for high-quality large-scale data collection with the human teleoperator in mind. By systematically analyzing different teleoperation frameworks and feedback interfaces, we aim to optimize data collection for scalable mobile manipulation robot learning.

\section{Related Work}
User interface design is a critical component of teleoperation, particularly for mobile manipulation. Mobile ALOHA~\cite{fu2024mobilealohalearningbimanual} converted the ALOHA setup~\cite{zhao2023learning} into a mobile system in which a replica of the bimanual robotic setup is used for teleoperation. Their findings indicate that naive teleoperators achieved near-expert performance after five trials, emphasizing the learning curve associated with well-designed interfaces. However, replicating a robotic setup is a difficult task. The TeleMoMa system \cite{dass2024telemoma} explored different input modalities, including VR controllers, vision-based tracking, and space mice, to assess their impact on user performance. Results indicated that hybrid approaches, where multiple control schemes are available, improved both intuitiveness and precision. Zhao et al. \cite{zhao2023learning} studied how teleoperation frequency affects performance, demonstrating that reducing control frequency from 50Hz to 5Hz led to a 62\% increase in task completion time. 

Recent studies have examined multimodal control systems that integrate vision, haptics, and auditory feedback to enhance teleoperation effectiveness~\cite{
Integrating_and_evaluating_visuo_tactile_sensing_with_haptic_feedback_for_teleoperated_robot_manipulation, 
patel2022haptic,
bong2022force, 
lippi2024low, 
sharma2023intuitive, 
patzold2023audio, 
su2023integrating,
wang2024robotic, 
garcia2022augmented, 
galarza2023virtual, 
su2022mixed, 
dass2024telemoma,
}.
In~\cite{lemasurier2022designing} and \cite{LeMasurier2024} the authors compared VR and traditional 2D interfaces, finding that VR offered better spatial awareness at the cost of longer task completion times. Similarly, exoskeleton-based teleoperation setups have shown good performance in teleoperation tasks~\cite{zhao2023wearable}. 
Learning-based teleoperation has been applied to humanoids using Mixed Reality~\cite{penco2024mixedrealityteleoperationassistance} and 3D Human Pose Estimation~\cite{he2024learninghumantohumanoidrealtimewholebody}.

Overall, advancements in teleoperation interfaces have focused on improving ergonomics, reducing workload, and increasing task success rates through multimodal interaction and adaptive control strategies. In this work, we provide useful insights along these lines on the effectiveness and ease of use for teleoperation in cognitively challenging long-horizon mobile manipulation scenarios by studying the effects of using different control embodiments and feedback modalities on the teleoperators' experience.


\begin{figure*}[t!]
    \centering
    \vspace{1em}
    \begin{subfigure}[b]{0.3\textwidth}
        \includegraphics[width=\textwidth]{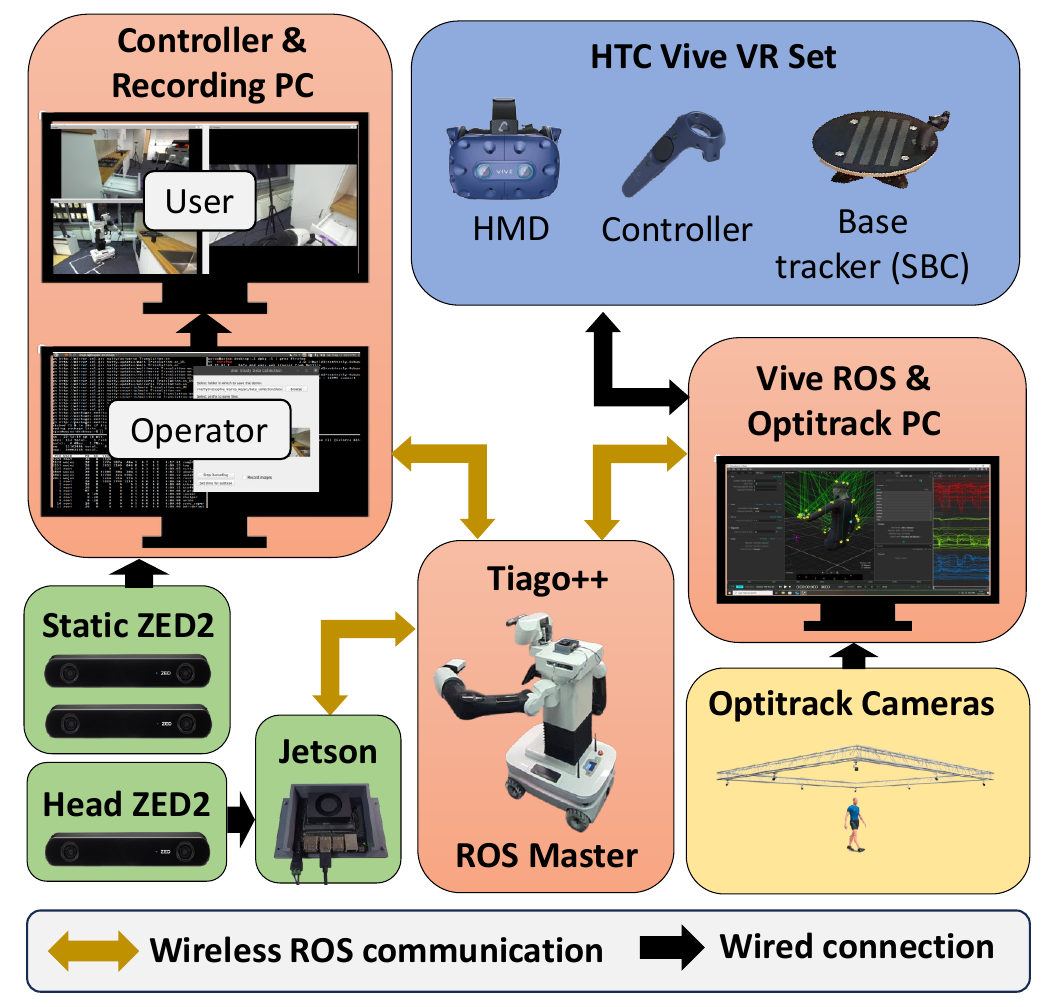} 
        \caption{}
        \label{fig:subfig1}
    \end{subfigure}
    \hfill
    \begin{subfigure}[b]{0.3\textwidth}
        \includegraphics[width=\textwidth]{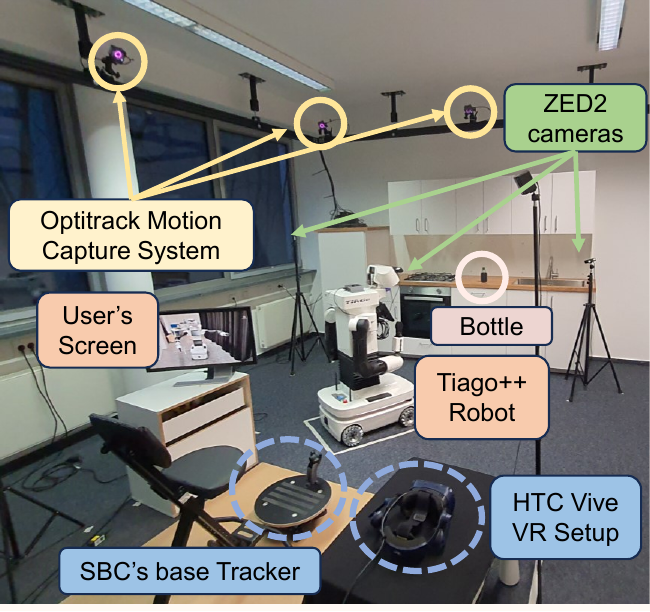} 
        \caption{}
        \label{fig:2d-map}
    \end{subfigure}
    \begin{subfigure}[b]{0.3\textwidth}
    \includegraphics[width=\textwidth]{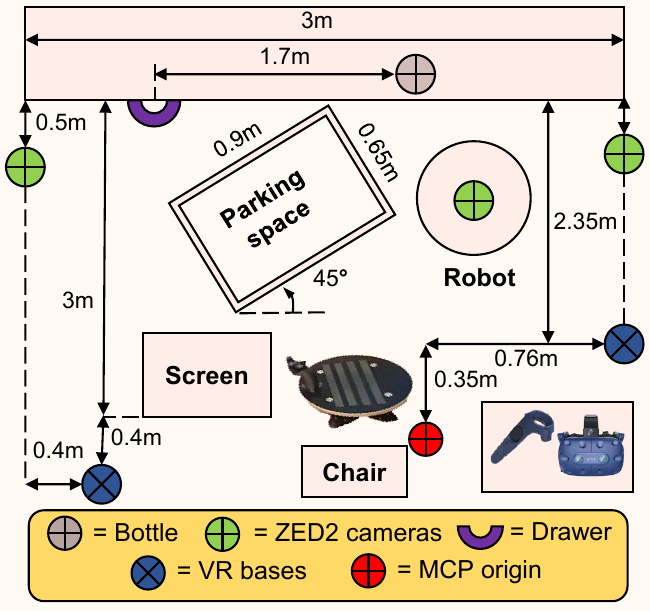} 
    \caption{}
    \label{fig:subfig2}
    \end{subfigure}
    \hfill
    \vspace{-0.5em}
    \caption{Detailed overview of \textbf{(a)} the components of robot teleoperation, data recording and visual feedback, and how they communicate, \textbf{(b)} their usage in the teleoperation setup, and \textbf{(c)} the room arrangement used for the real-world task sequence.
    }
    \label{fig:overview_setup}
\end{figure*}

\section{Teleoperation System Design}
In this paper, we study how different control embodiments and feedback modalities in a teleoperation setup affect users' ability to comfortably and efficiently perform complex tasks that require both navigation and precise object manipulation.
The user study objectively compares different system interfaces through a tailored task sequence. In addition to task performance, we collect a range of data on cognitive and physical workload, along with user experience.

The teleoperation setup is composed of an HTC Vive Pro Virtual Reality (VR) setup to control a PAL Tiago++ robot with an omnidirectional base. This user study compares two different \textit{Controllers}: a \textbf{W}hole \textbf{B}ody \textbf{C}ontroller (WBC) and  a \textbf{S}eparate \textbf{B}ody \textbf{C}ontroller (SBC). For providing feedback to the teleoperator, we study two distinct \textit{Modalities}: with Virtual Reality (i.e., the VR headset) or without VR (on an external screen). In total, there are four possible combinations of interfaces. Both controllers leverage a joint impedance controller, which enables safe interaction. The entire setup with the different control and feedback modalities is shown in Fig.~\ref{fig:overview_setup}. A simulation environment in Gazebo is built replicating the real study scenario and is used for training purposes only.

\subsection{Controller Embodiments}

The \textbf{SBC} controller consists of independent control systems that decouple the base motion from the arm motion. This decoupling provides an operator the option to separately control either embodiment as required.
Following \cite{flacco_2015_iknull} and \cite{Zhu2023}, the arm controller employs an inverse kinematics (IK) solver with null-space resolution to compute the desired joint angle configurations based on the relative change in the end-effector pose estimated from VR controllers at 30 Hz. 
For the null-space optimization, we use a manipulability criterion to favor feasible arm postures and avoid unreachable task poses, similar to \cite{nakanishi_2020_iknull_manipulability, Zhu2023,nakanishi_2008_iknull_manipulability}.  
The IK solver is implemented using the Pinocchio motion library \cite{carpentier_2019_pinocchio}.
The mobile base employs a 3D rudder with an attached VR tracker, inspired by \cite{Lenz2023}, to translate velocity inputs into relative position changes of the base.
The \textbf{WBC} framework~\cite{Lueth2024} running at 15~Hz combines a whole body controller to compute desired joint-space motion via Task Space Inverse Dynamics by QP, using the TSID library~\cite {delprete2015}, and joint impedance. The user, who is only using a VR controller, can switch between end effector (EE) mode and whole-body manipulation (WBM) mode, which trade off different sets of task objectives. Whereas in EE mode, direct teleoperation of the EE without base motion is prioritized, the WBM mode keeps the EE stable while the user directly teleoperates elbow and the base while considering self-collision avoidance.
\subsection{Feedback Modalities}

In the \textbf{With VR} modality, the participants were asked to wear the VR headset and were able to switch between 2 stereo cameras placed statically around the room and 1 stereo camera placed on top of the robot's head. The robot head movement was additionally controlled by the headset motion. In the \textbf{Without VR} modality, participants did not wear the headset and could look around the room while simultaneously viewing all three camera streams on a screen. 

\subsection{Task Sequence}
\label{ssec:task-sequence}
To assess the capabilities of the \textit{Controllers}, we asked users to perform tasks that require reasonable use of all navigation and object manipulation features applied to various tasks. To assess the capabilities of the visualization \textit{Modalities}, the proposed tasks should require refined pose estimation of objects and good localization of the robot itself by the user. We adopt, therefore, a long-horizon sequence of tasks in a kitchen environment. The robot starts in an angled parking space with the left arm in a raised pose. While seated, the user must control the robot with their left arm to perform the following sequence of tasks: Drive toward the drawer and open it (Task 1); pick up a bottle on the other side of the kitchen counter (Task 2); drop the bottle inside the drawer (Task 3); close the drawer (Task 4); and park the robot back in its parking space (Task 5). This proposal not only realistically attests the mobility capabilities of the system but can also significantly reduce the time needed to perform the user study since the tasks are sequential, with no need for complex environment resets for the next trial. A schematic of the experimental setup is shown in Fig.~\ref{fig:2d-map}.

Each task requires the effective coordination of both the upper and lower body of the robot to navigate to a given location, accurately move the arm to a target pose, and seamlessly synchronize the motion of both the upper and lower body to complete the task. For example, in Task 1, the operator must navigate the robot base so that the drawer is within the reach of the robot to grasp the handle while allowing enough space for the robot to open the drawer, demanding fine-tuned spatial reasoning and control. Task 2 introduces additional complexity, as the operator must navigate the robot across the kitchen and precisely position the arm to grasp a bottle on the counter, requiring careful coordination between base mobility and arm dexterity. Task 3 further amplifies the challenge, as the operator must maintain stability while transporting the bottle and align the arm to place it inside the drawer, requiring precise timing and spatial awareness. Task 4 involves closing the drawer, which requires delicate force control and alignment to avoid collisions, while Task 5 demands accurate navigation and positioning to return the robot to its parking space, often under time constraints. 

Moreover, given the long-horizon nature of the task sequence, operators are prone to fatigue, which may subsequently increase the cognitive load to effectively coordinate both the upper and lower body of the robot. The need for continuous attention to both navigation and manipulation, coupled with precise pose estimation and localization, underscores the complexity of teleoperation for mobile manipulation. This challenge is further exacerbated by the sequential dependency of tasks, where errors in early stages can compound, making recovery difficult and increasing the overall difficulty of the operation.

\section{User Study}

\subsection{Study Design}
We test 2 independent variables as part of our study: \textit{Controller} and visualization \textit{Modality}. We test 2 different controllers: \textbf{SBC} and \textbf{WBC}. For the \textit{Modality}, we compare the use of the VR Headset (``with VR") or an external display (``without VR"). In total, we have 4 combinations of user interfaces. We test each combination 3 times to track the improvement of the user performance over runs of the proposed tasks. This adds another dimension to the independent variables of the study that we will call \textit{Trial}. 

Similar to LeMasurier et al. \cite{LeMasurier2024}, we opt for a 2 (\textit{Controller}) $\times$ 2 (\textit{Modality}) $\times$ 3 (\textit{Trial}) mixed study design, with \textit{Controllers} being tested between-subjects and \textit{Modalities} and \textit{Trials} within-subjects. An Overview can be found in Table~{\ref{tab:modality_controller}}. Having $N$ registered participants, each half $\sfrac{N}{2}$ tests only one of the two \textit{Controllers}, which requires all registered participants to be stratified according to personal features that may influence the outcome of the study (e.g. experience with VR). To track the learning curve, the participants perform 3 \textit{Trials} of the assigned controller with both \textit{Modalities}. To mitigate priming bias, we randomize the order in which each \textit{Modality} will be tested by the participant. In total, each participant would then do 6 runs of the task sequence (i.e. 3 Trials with each of the 2 Modalities).

\begin{table}[h!]
\centering
\begin{tabular}{|l||l|l|}
\cline{2-3}
\multicolumn{1}{c|}{} & \textbf{Type of Analysis} & \textbf{Options} \\ 
\hline
\textbf{Controller} & Between-Subjects & WBC / SBC \\ \hline 
\textbf{Modality} & Within-Subjects & With VR / Without VR \\ \hline
\textbf{Trial} & Within-Subjects & 1 / 2 / 3 \\ \hline
\end{tabular}
\caption{Options and type of analysis for each of the controlled variables.}
\label{tab:modality_controller}
\end{table}

\subsection{User Study Protocol}
The whole study for each participant lasts around 2 hours. Initially, participants were asked to fill out a personal data questionnaire with relevant questions for the stratification between \textit{Controllers} as well as consent forms. We then asked the participant to wear an upper-body motion-tracking suit, following which a calibration procedure was done for the motion capture to accurately track the participant's body. After the calibration, the order of the \textit{Modality} is randomly selected, and a short instruction about the system and the task is given. Once the participant understood the system and the task, they were given 6 minutes to train in simulation with each \textit{Modality} in the defined order. The participant was initially given the choice of freely controlling the robot without any task at hand. Once they felt confident, they were asked to attempt the first task of reaching and opening the drawer, and subsequently of grasping the bottle. This was done to allow participants to get used to the system. After the simulation training phase with both modalities (with and without VR), the participant was then given 4 minutes of training time in the real world, similar to the VR training phase according to the order of the modalities. Once the real-world training was done for a given modality, the participant was then asked to teleoperate the robot to perform the tasks described in Sec.~\ref {ssec:task-sequence}. The participants had 3 trials in the real world to perform the overall task sequence. Once the 3 trials for the first modality were done, the entire process of real-world training followed by the 3 trials for the task sequence was repeated for the second modality. 

\subsection{Metrics}
\label{subsec:workload}

To evaluate our user study, we use a combination of behavioral and attitudinal metrics. Behavioral metrics include ergonomics data, robot data, VR setup data, and task performance data, such as completion times and performance scores (e.g., Success: 10, Partial Success: 7, Partial Failure: 4, Failure: 0). We also track motion data using an Optitrack system to calculate postural scores (RULA) and Center of Mass (CoM) divergence~\cite{Gholami2021quantitative} for the left upper arm. 
Moreover, all information related to the HTC Vive controller, headset, and tracker poses and velocities as well as the robot's joint states and chosen camera stream are recorded in a ROSBAG.

Attitudinal metrics are gathered through standardized questionnaires, simplified to have questions relevant to our scenario, and administered at different stages of the study. We do so to reduce participant fatigue during the experiment.  Short usability (SEQ)~\cite{seq} and workload (Air Force Flight Test Center Revised Workload Estimate Scale ``ARWES/CSS"~\cite{arwes}) questionnaires are collected after each trial. More detailed assessments, like the NASA Task Load Index (TLX)~\cite{moroney1992comparison,hart1988nasa} for workload, the Usability Metric for User Experience (UMUX)~\cite{finstad2011umux} for usability, and the Operational Assessment of Training Scale (OATS)~\cite{oats} for training effectiveness, are conducted after all three trials for a given feedback \textit{Modality}. Additionally, a simplified version of the Simulation Sickness Questionnaire (SSQ) ~\cite{kennedy1993ssq,singla_2021_ssq_reduced} is used to assess discomfort after the ``with VR" Modality trials. Lastly, participants provided feedback on the interface comparisons upon completing the study.

\section{Results}
The study was conducted for 20 participants, stratified as equally as possible between both controllers according to selected relevant features, including \mbox{VR-,} videogame-, teleoperation- and driving-experience as well as handedness, gender and possession of eyesight conditions. Participants were mostly young adults (SBC: 24.4y$\pm$3.9, WBC: 25.4y$\pm$3.2) holding or pursuing a Master's degree (SBC: 60\%, WBC: 70\%) and right-handed (SBC: 9, WBC: 9) working in the field of engineering (90\% overall).

Most metrics collected failed the Shapiro-Wilk normality test (p$<$0.05), leading to the usage of non-parametric tests for statistical significance verification. For metrics collected after all trials of each \textit{Modality}, a Mann-Whitney U Test was applied to the between-subject variable \textit{Controller} and a Wilcoxon Signed-Rank Test was applied for the within-subject variable \textit{Modality}. For metrics collected every trial, Linear Mixed-Effects Model (LMM) was, due to the added dimension of the \textit{Trials} effect, representing repeated measures. LMM includes random effects at the participant level to control for individual differences, making it more robust than repeated-measures ANOVA, which assumes sphericity and normality. The results of the statistical tests can be seen in Table \ref{tab:hypothesis_results} and will be explained in further detail in the following subsections.

\begin{table}[h]
    \centering
    \renewcommand{\arraystretch}{1.2} 
    \begin{tabular}{|lccc|}
        \hline
        \textbf{Assessment} & \textbf{Controller} & \textbf{Modality} & \textbf{Trial} \\
        \hline
        \multicolumn{4}{|c|}{\textbf{Usability}} \\
        \hline
        SEQ & None & Strong & Marginal \\
        UMUX & None & Strong & NA \\
        \hline
        \multicolumn{4}{|c|}{\textbf{Workload}} \\
        \hline
        ARWES & None & Very strong & None \\
        Physical Demand & Slight & Strong & N/A \\
        Mental Demand & None & Strong & N/A \\
        Temporal Demand & None & Marginal & N/A \\
        Performance & None & Strong & N/A \\
        Frustration & None & Slight & N/A \\
        Effort & None & Very strong & N/A \\
        \hline
        \multicolumn{4}{|c|}{\textbf{Ergonomics}} \\
        \hline
        RULA & None & None & None \\
        \hline
        \multicolumn{4}{|c|}{\textbf{Task Performance}} \\
        \hline
        Completion Times & Slight & Slight & None \\
        Success Rate & None & None & None \\
        \hline
    \end{tabular}
    \caption{Statistical test results on the impact of Controller, Modality and Trial on Human Workload and Task Performance indicated by p-value. None $p>0.1$, Marginal $0.1 > p > 0.05$, Slight $0.05 > p > 0.01$, Strong $0.01 > p > 0.001$ and Very Strong $p < 0.001$.}
    \label{tab:hypothesis_results}

\end{table}

\subsection{Task Performance}
\subsubsection{Completion Times}
The choice of \textit{Modality} and \textit{Controller} has a significant impact on task completion time. The usage of VR increases total completion time by 142 seconds (p=0.026) due to limited depth perception. The SBC Controller with its separate base movement leads to a faster task completion time (-169 seconds,  p=0.025) compared to the WBC Controller. The number of \textit{Trials} had a positive effect on completion time (-31.64 seconds per trial, p=0.12), indicating a learning effect over repeated attempts.

\subsubsection{Success Rate}
Participants consistently demonstrated a high level of task execution success across all conditions, with an average task score of 9.4 out of 10 (p$<$0.0001). Controller type and visualization modality showed no statistically significant effect on success rate. WBC (-1.50 points, p=0.27) and VR (-1.20 points, p=0.37) performed slightly worse on average. Performance remained stable across \textit{Trials}, with no notable learning or fatigue effects. Furthermore, there was no significant variation across different tasks, meaning no single task was consistently harder or easier than the others. Unlike the completion time analysis, where \textbf{with VR} significantly increased task duration, here we see that performance scores remained unaffected by visualization modality, implying that while VR may slow down execution, it does not necessarily lead to task failure or lower performance quality.

\begin{figure}[t!]
\vspace{1em}
    \centering
    \includegraphics[width=\columnwidth]{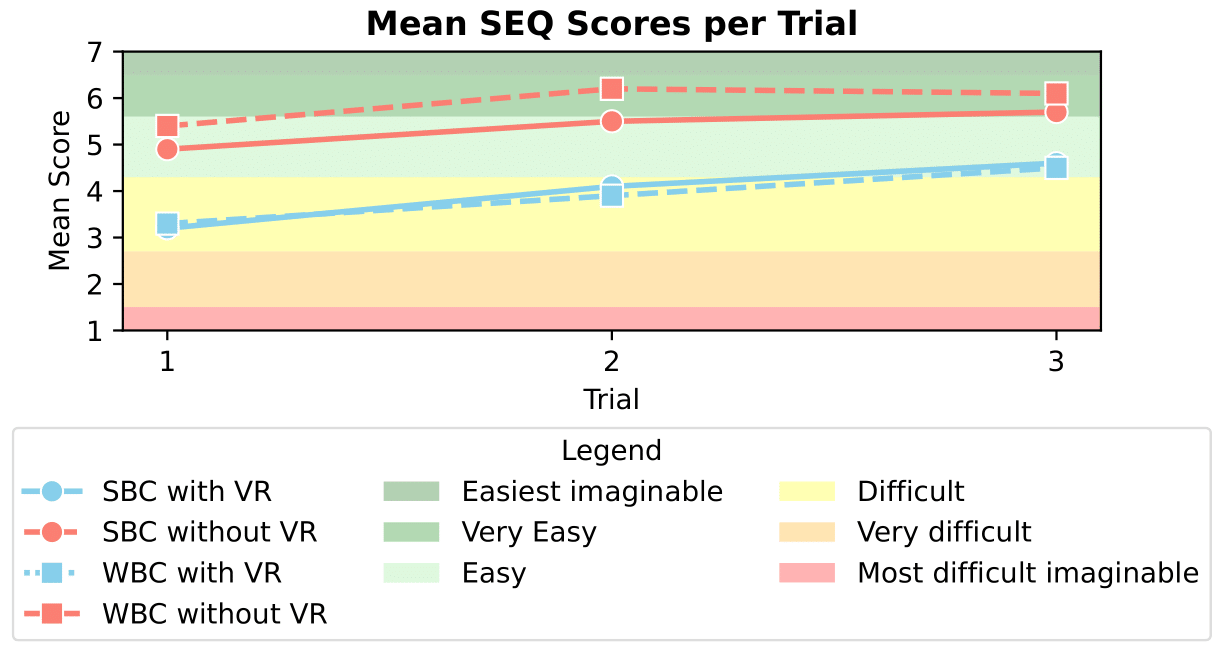}
    \caption{Mean values for the Simple of Ease Question (SEQ) over trials across controllers and modalities in the real world experiments. A higher SEQ score indicates better usability and ease.}
    \label{fig:seq_trials}
\end{figure}

\subsection{Human Interface Assessment} 

\subsubsection{Usability}
The \textbf{SEQ} questionnaire that was collected over all \textit{Trials} shows only the easiness of use dimension of usability, while the \textbf{UMUX} was collected only after the end of each \textit{Modality} test and gives a more comprehensive overview of the usability scope. The main effect of trial number suggests a marginal but not statistically significant increase in \textbf{SEQ} scores over trials (p=0.068). This indicates that there may be a slight learning effect, where participants find it easier to accomplish the proposed task over repeated trials. The main effect of \textit{Controllers} was not significant for the \textbf{SEQ} scores (p=0.386). This suggests that the mean \textbf{SEQ} scores for \textbf{SBC} and \textbf{WBC} cannot be statistically differentiated, indicating similar post-trial interface complexity perception. Detailed results can be seen in Figure~\ref{fig:seq_trials}.
However, when looking at the \textbf{UMUX} results, a suggestively better usability score for the \textbf{SBC} over the \textbf{WBC} (p=0.15) was found, indicating that other dimensions of usability not included in \textbf{SEQ}, such as frustration, expectation fulfillment or frequent error compensation, may be the main disadvantage of the \textbf{WBC} over the \textbf{SBC}. Regarding the \textit{Modalities}, participants found tasks significantly harder in the \textbf{with VR} condition compared to \textbf{without VR} according to both usability scores \textbf{SEQ} (p=0.003) and \textbf{UMUX} (p=0.006).

\begin{figure}[t]
    \centering
    \includegraphics[width=1.0\columnwidth]{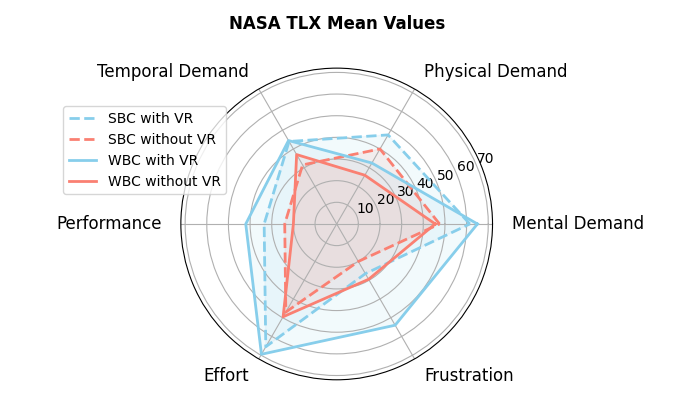}
    \caption{Radar plot of the mean values of the NASA TLX individual scores for all 4 interface combinations. The scale goes from 0 (not demanding) to 100 (extremely demanding). A smaller area corresponds to a lower workload for the user. Both combinations with VR (in blue) appear to have a significant higher workload across all features. For almost all features, the WBC has worse results than the SBC, except for "Physical Demand", which is slightly higher for the SBC.}
    \label{fig:nasa-tlx}
    \vspace{-0.5cm}
\end{figure}

\subsubsection{Workload}
\begin{figure}[h]
    \centering
    \includegraphics[width=0.8\columnwidth]{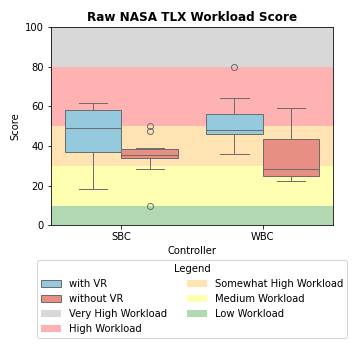}
    \caption{Boxplot of the overall raw NASA TLX score. An interpretation benchmark for the NASA TLX results is shown, associating the scores level of workload, as defined in \cite{nasa_tlx_benchmark}. The average scores for with VR modality are higher than without VR, falling in the high workload classification, while the average score for the trials without VR is considered somewhat high workload.}
    \label{fig:nasa_boxplots}
    \vspace{-0.5cm}
\end{figure}

The \textbf{ARWES} questionnaire was collected over all \textit{Trials} and resumes the workload assessment with only one question, while the \textbf{NASA TLX} was collected only after the end of each \textit{Modality} test and gives a more comprehensive overview of the cognitive- and physical demands of the system. \textbf{ARWES} results indicate that the usage of the VR HMD imposed a substantially higher workload on the user than the usage of assistive screens. Results of the \textbf{NASA TLX} confirm this, indicating the \textbf{with VR} \textit{Modality} resulted in higher perceived workload, requiring more cognitive and physical effort while reducing performance. The statistical testing and mean TLX scores also suggest that while the type of \textit{Controller} had minimal influence on other workload dimensions, \textbf{SBC} induced more perceived physical strain (p=0.02), whereas \textbf{WBC} led to higher frustration levels (p=0.009), which aligns with the discussion presented in the usability results section. For more detailed results of the NASA TLX see Figure \ref{fig:nasa-tlx} and \ref{fig:nasa_boxplots}.

\subsubsection{Ergonomics}
\begin{figure}

\centering
\vspace{1em}
\begin{subfigure}{1.0\columnwidth}
  \centering
  \includegraphics[width=0.975\columnwidth]{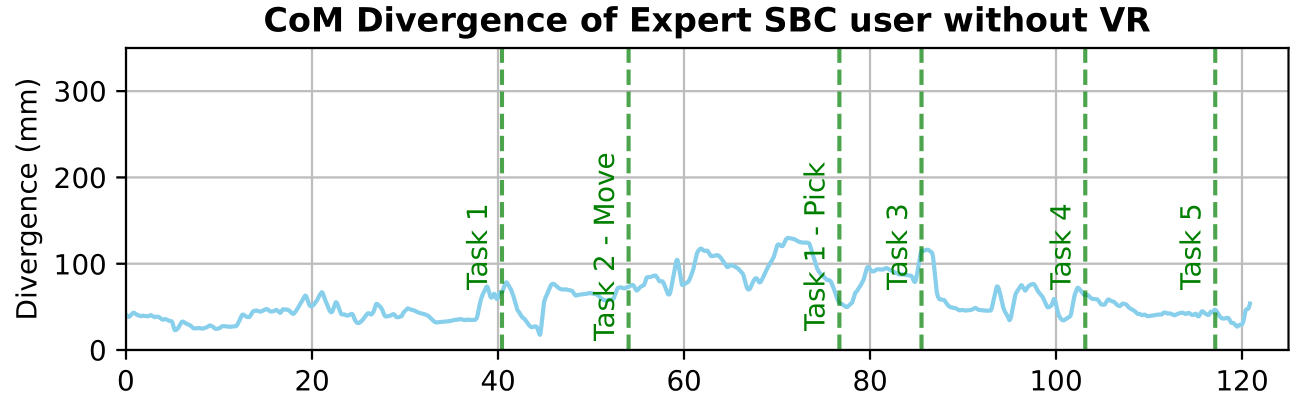}
\end{subfigure}%

\begin{subfigure}{1.0\columnwidth}
  \centering
  \includegraphics[width=1.0\columnwidth]{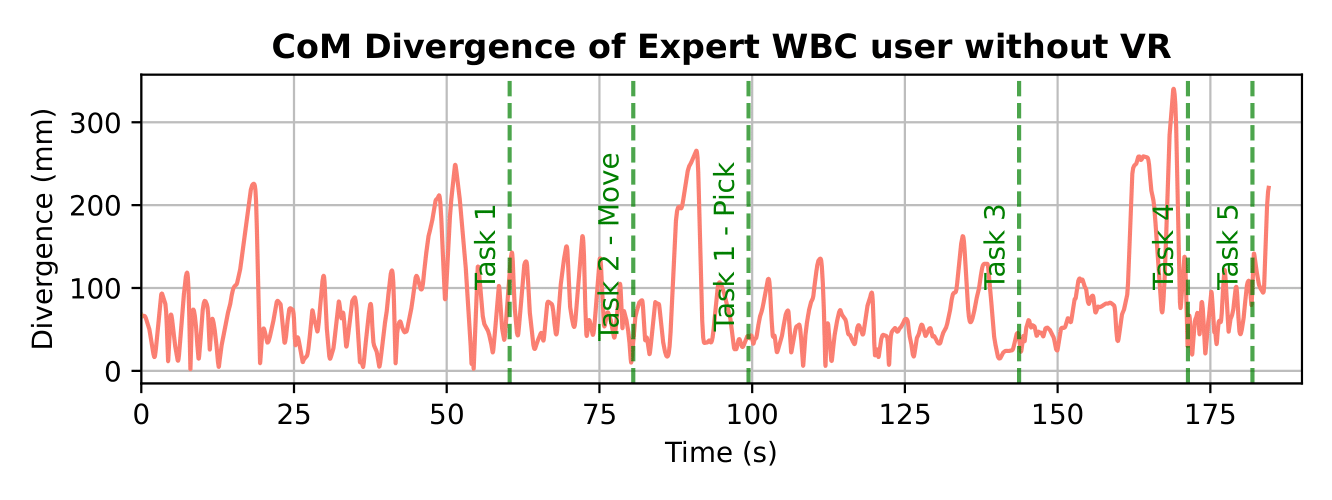}
\end{subfigure}
\caption{Comparison of CoM Divergence of both controllers without VR from expert users shows frequent CoM shifts for \textbf{WBC}.}
\label{fig:expert_com}
\end{figure}

Regarding the final \textbf{RULA} score for interface combination comparison, the results showed no significant effect of \textit{Controller} (p=0.6) or \textit{Modality} (p=0.4), which suggests that ergonomic risk did not vary significantly between \textbf{SBC} and \textbf{WBC}, nor between \textbf{with VR} and \textbf{without VR} conditions. Similarly, there was no significant effect of \textit{Trial} number (p=0.29), indicating that participants did not experience substantial ergonomic improvements or declines over repeated trials. The main goal here was, however, to evaluate the proposed teleoperation setup in terms of a global ergonomic benchmark. The mean total RULA score was 
4.12$\pm$0.27, 
indicating medium musculoskeletal disorders risk over prolonged sessions. The higher RULA score is mainly to the upper arm subscore of RULA that had a mean of 
3.28$\pm$0.23,
indicating average operation of the upper arm at an elevation angle between 45° and 90°. The wrist subscore was also relatively high, with 
2.86$\pm$0.22,
indicating frequent wrist bend  of more than 15°, essential for teleoperating end-effector (EE) movements for both \textit{Controllers}. The scores for neck, trunk and lower arm are between 1 and 2, the lowest possible, indicating mostly upright position and horizontal forearm pose. Furthermore, we can look at the \textbf{CoM divergence} results over time for signs of excessive whole-body engagement and postural instability. By analyzing an example of the task being performed by expert users in Figure~\ref{fig:expert_com}, we observe that the WBC exhibits significantly greater variation in CoM divergence values than SBC. In the WBC, for locomotion, the base motion is temporarily activated by the user, whose controller's pose difference is then mapped the robot's wheels velocities. Thus, the increased CoM divergence is inherent to the design of the WBC control strategy for integrating locomotion, rather than a result of user inexperience. The greater variations in CoM for WBC indicate a more physically demanding control method, which may contribute to greater fatigue over extended usage.

\vspace{0.5cm}  

\subsection{Virtual Reality}
In the \textbf{with VR} \textit{Modality}, users of the \textbf{SBC} used more the head camera view than users of the \textbf{WBC} (t-test:~p$<$0.0001), namely $\mu$=60.4$\pm$38\% of the task trials for \textbf{SBC} and $\mu$=36.8$\pm$39\% for the \textbf{WBC}, indicating a higher sense of embodiment and confidence for \textbf{SBC} uses. For both \textit{Controllers}, Tasks 3 and 5 had a higher usage of the 3rd-person view cameras for locomotion assistance. As to the analysis of the 9-symptoms \textbf{SSQ} questionnaire, a correction factor allowed us make an approximate interpretation of from the 16-symptoms SSQ benchmark. Results indicate the experimented VR sickness for the real world experiments is on the edge of significance to concerning. For the simulation trials, VR sickness could be considered minimal, which is also a reflection of shorter amount of time the users spent teleoperating in simulation. Identified improvements in the video stream delays and video resolution could substantially reduce the experimented VR sickness. 

\subsection{Simulation Training}
The reduced \textbf{OATS} results indicate that the simulation training is of relatively high relevance ($\mu$=4.75$\pm$1.2), efficacy ($\mu$=4.8$\pm$1.2) and overall quality ($\mu$=4.78$\pm$1.2) on a 7-point Likert scale. \textbf{SEQ} ease of use results indicate completion of the tasks in simulation more difficult (\textbf{with VR}: p=0.015, \textbf{witout VR}: p=$<$0.0001) than in real world, but not significantly more physically or mentally demanding according to \textbf{ARWES} results (p$>$0.8). The higher perception of task difficulty is a reflection not only of the priming bias, since simulation was the participants' first interaction with the system, but also reportedly lower sense of presence due to lack of audio feedback when compared to teleoperating the real robot in the same room.

\section{Preliminary Imitation Learning Experiment}

We conducted a small-scale experiment to assess the suitability of data collected with SBC and WBC for downstream imitation learning.
We collected 50 trajectories for each controller without VR for the first task of reaching and opening a drawer, and trained a transformer-based Diffusion Policy~\cite{chi2024diffusionpolicy} without visual input.
The state space 
consists of the robot's 10-DoF joint state (3 for the base and 7 for the arm), end-effector 3D position and 1D position of the drawer opening acquired by Optitrack. The action space consists of the 10 DoF desired joint state and the gripper state.

We report success rates of 0~\% and 80~\% out of 5 trials with the SBC and WBC data, respectively. 
We attribute the failure of the SBC policy due to the lack of base-arm coupling in the motion signals which make the learned policy more susceptible to encounter out-of-distribution states.
While not statistically representative, we take this as an indication that WBC data is better suited for our imitation learning approach.
\section{Conclusion and Future Work}
In this article, we present a comprehensive user study on the usability of teleoperation interfaces for mobile manipulation. We assess multiple combinations of embodiment and visual feedback on a long-horizon mobile manipulation task sequence.
Our study indicates that while both the coupled and decoupled embodiment of manipulation and navigation lead to comparable workload on the user, they induce different strategies for solving the task.
Our SBC, which decouples arm and base control interfaces, achieves shorter completion times by users exploiting the direct base control and less frustration. Furthermore, our results show that visual feedback in the form of VR instead of screen-based camera streams increases cognitive and physical workload. 
A thorough investigation of the collected data quality for imitation learning remains to be done; however, preliminary results indicate better data quality in motion data that couples arm and base motion, generated by the WBC. Therefore, we plan to enhance usability of our WBC  with simplified base control for future data collection.
Also, an analysis of the effects of extending the teleoperation controllers to different feedback modalities like haptic or audio remains a promising extension of this work.



\printbibliography

\end{document}